\documentstyle[latex-acl]{article}

\newcommand{\ifshort}[1]{}
\newcommand{\iflong}[1]{#1}

\newcommand{\shortfootnote}[1]{\ifshort{\footnote{#1}}}
\newcommand{\longfootnote}[1]{\iflong{\footnote{#1}}}

\title{\vspace{-0.5in}Improving Language Models by \\ Clustering Training
Sentences}

\author{David Carter \\ SRI International \\ 23 Millers Yard \\
Cambridge CB2 1RQ, UK \\ \verb!dmc@cam.sri.com!}

\ifshort{}

\begin{document}

\maketitle
\vspace{-0.5in}
\begin{abstract}

Many of the kinds of language model used in speech understanding
suffer from imperfect modeling of intra-sentential contextual
influences. I argue that this problem can be addressed by clustering
the sentences in a training corpus automatically into subcorpora on
the criterion of entropy reduction, and calculating separate language
model parameters for each cluster.  This kind of clustering offers a
way to represent important contextual effects and can therefore
significantly improve the performance of a model. It also offers a
reasonably automatic means to gather evidence on whether a more
complex, context-sensitive model using the same general kind of
linguistic information is likely to reward the effort that would be
required to develop it: if clustering improves the performance of a
model, this proves the existence of further context dependencies, not
exploited by the unclustered model. As evidence for these claims, I
present results showing that clustering improves some models but not
others for the ATIS domain. These results are consistent with other
findings for such models, suggesting that the existence or otherwise
of an improvement brought about by clustering is indeed a good pointer
to whether it is worth developing further the unclustered model.

\end{abstract}

\section{1. Introduction}

In speech recognition and understanding systems, many kinds of
language model may be used to choose between the word and sentence
hypotheses for which there is evidence in the acoustic data.  Some
words, word sequences, syntactic constructions and semantic structures
are more likely to occur than others, and the presence of more likely
objects in a sentence hypothesis is evidence for the correctness of
that hypothesis. Evidence from different knowledge sources can be
combined in an attempt to optimize the selection of correct hypotheses;
see e.g.\ Alshawi and Carter (1994); Rayner {\it et al} (1994);
Rosenfeld (1994).

Many of the knowledge sources used for this purpose score a sentence
hypothesis by calculating a simple, typically linear, combination of
scores associated with objects, such as $N$-grams and grammar rules,
that characterize the hypothesis or its preferred linguistic analysis.
When these scores are viewed as log probabilities, taking a linear sum
corresponds to making an independence assumption that is known to be
at best only approximately true, and that may give rise to
inaccuracies that reduce the effectiveness of the knowledge source.

The most obvious way to make a knowledge source more accurate is to
increase the amount of structure or context that it takes account of.
For example, a bigram model may be replaced by a trigram one, and the
fact that dependencies exist among the likelihoods of occurrence of
grammar rules at different locations in a parse tree can be modeled
by associating probabilities with states in a parsing table rather
than simply with the rules themselves (Briscoe and Carroll, 1993).

However, such remedies have their drawbacks. Firstly, even when the
context is extended, some important influences may still not be
modeled. For example, dependencies between words exist at separations
greater than those allowed for by trigrams (for which long-distance
$N$-grams [Jelinek {\it et al}, 1991] are a partial remedy), and
associating scores with parsing table states may not model all the
important correlations between grammar rules. Secondly, extending the
model may greatly increase the amount of training data required if
sparseness problems are to be kept under control, and additional data
may be unavailable or expensive to collect. Thirdly, one cannot always
know in advance of doing the work whether extending a model in a
particular direction will, in practice, improve results. If it turns
out not to, considerable ingenuity and effort may have been wasted.

In this paper, I argue for a general method for extending the
context-sensitivity of {\it any} knowledge source that calculates
sentence hypothesis scores as linear combinations of scores for
objects. The method, which is related to that of Iyer, Ostendorf and
Rohlicek (1994), involves clustering the sentences in the training
corpus into a number of subcorpora, each predicting a different
probability distribution for linguistic objects. An utterance
hypothesis encountered at run time is then treated as if it had been
selected from the subpopulation of sentences represented by one of
these subcorpora.  This technique addresses as follows the three
drawbacks just alluded to.  Firstly, it is able to capture the most
important sentence-internal contextual effects regardless of the
complexity of the probabilistic dependencies between the objects
involved.  Secondly, it makes only modest additional demands on
training data.  Thirdly, it can be applied in a standard way across
knowledge sources for very different kinds of object, and if it does
improve on the unclustered model this constitutes proof that
additional, as yet unexploited relationships exist between linguistic
objects of the type the model is based on, and that therefore it is
worth looking for a more specific, more powerful way to model them.

The use of corpus clustering often does not boost the power of the
knowledge source as much as a specific hand-coded extension. For
example, a clustered bigram model will probably not be as powerful as
a trigram model. However, clustering can have two important uses. One
is that it can provide some improvement to a model even in the absence
of the additional (human or computational) resources required by a
hand-coded extension. The other use is that the existence or otherwise
of an improvement brought about by clustering can be a good indicator
of whether additional performance can in fact be gained by extending
the model by hand without further data collection, with the possibly
considerable additional effort that extension would entail. And, of
course, there is no reason why clustering should not, where it gives
an advantage, also be used in conjunction with extension by hand to
produce yet further improvements.

As evidence for these claims, I present experimental results showing
how, for a particular task and training corpus, clustering produces a
sizeable improvement in unigram- and bigram-based models, but not in
trigram-based ones; this is consistent with experience in the speech
understanding community that while moving from bigrams to trigrams
usually produces a definite payoff, a move from trigrams to 4-grams
yields less clear benefits for the domain in question. I also show
that, for the same task and corpus, clustering produces improvements
when sentences are assessed not according to the words they contain
but according to the syntax rules used in their best parse. This work
thus goes beyond that of Iyer {\it et al} by focusing on the
methodological importance of corpus clustering, rather than just its
usefulness in improving overall system performance, and by exploring
in detail the way its effectiveness varies along the dimensions of
language model type, language model complexity, and number of clusters
used. It also differs from Iyer {\it et al}'s work by clustering at
the utterance rather than the paragraph level, and by using a training
corpus of thousands, rather than millions, of sentences; in many
speech applications, available training data is likely to be quite
limited, and may not always be chunked into paragraphs.

\section{2. Cluster-based Language Modeling}

Most other work on clustering for language modeling (e.g.\ Pereira,
Tishby and Lee, 1993; Ney, Essen and Kneser, 1994) has addressed the
problem of data sparseness by clustering {\it words} into classes
which are then used to predict smoothed probabilities of occurrence
for events which may seldom or never have been observed during
training. Thus conceptually at least,
their processes are agglomerative: a large initial set of words is
clumped into a smaller number of clusters. The approach described here
is quite different. Firstly, it involves clustering whole sentences,
not words.  Secondly, its aim is not to tackle data sparseness by
grouping a large number of objects into a smaller number of classes,
but to increase the precision of the model by dividing a single object
(the training corpus) into some larger number of sub-objects (the
clusters of sentences). There is no reason why clustering sentences
for prediction should not be combined with clustering words to reduce
sparseness; the two operations are orthogonal.

Our type of clustering, then, is based on the assumption that the
utterances to be modeled, as sampled in a training corpus, fall more
or less naturally into some number of clusters so that words or
other objects associated with utterances have probability distributions
that differ between clusters. Thus rather than estimating the relative
likelihood of an utterance interpretation simply by combining fixed
probabilities associated with its various characteristics, we view
these probabilities as conditioned by the initial choice of a cluster
or subpopulation from which the utterance is to be drawn. In both
cases, many independence assumptions that are known to be at best
reasonable approximations will have to be made.  However, if the
clustering reflects significant dependencies, some of the worst
inaccuracies of these assumptions may be reduced, and system
performance may improve as a result.

Some domains and tasks lend themselves more obviously to a clustering
approach than others.  An obvious and trivial case where clustering is
likely to be useful is a speech understander for use by travelers in
an international airport; here, an utterance will typically consist of
words from one, and only one, natural language, and clusters for
different languages will be totally dissimilar. However, clustering
may also give us significant leverage in monolingual cases. If the
dialogue handling capabilities of a system are relatively rigid, the
system may only ask the user a small number of different questions
(modulo the filling of slots with different values). For example, the
CLARE interface to the Autoroute PC package (Lewin {\it et al}, 1993)
has a fairly simple dialogue model which allows it to ask only a dozen
or so different types of question of the user. A Wizard of Oz
exercise, carried out to collect data for this task, was conducted in
a similarly rigid way; thus it is straightforward to divide the
training corpus into clusters, one cluster for utterances immediately
following each kind of system query. Other corpora, such as Wall
Street Journal articles, might also be expected to fall naturally into
clusters for different subject areas, and indeed Iyer {\it et al}
(1994) report positive results from corpus clustering here.

For some applications, though, there is no obvious extrinsic basis for
dividing the training corpus into clusters. The ARPA air travel
information (ATIS) domain is an example.  Questions can mention
concepts such as places, times, dates, fares, meals, airlines, plane
types and ground transportation, but most utterances mention several
of these, and there are few obvious restrictions on which of them can
occur in the same utterance.  Dialogues between a human and an ATIS
database access system are therefore likely to be less clearly
structured than in the Autoroute case.

However, there is no reason why automatic clustering should not be
attempted even when there are no grounds to expect clearly distinct
underlying subpopulations to exist. Even a clustering that only partly
reflects the underlying variability of the data may give us more
accurate predictions of utterance likelihoods. Obviously, the more
clusters are assumed, the more likely it is that the increase in the
number of parameters to be estimated will lead to worsened rather than
improved performance. But this trade-off, and the effectiveness of
different clustering algorithms, can be monitored and optimized by
applying the resulting cluster-based language models to unseen test
data. In Section \ref{expt} below, I report results of such
experiments with ATIS data, which, for the reasons given above, would
at first sight seem relatively unlikely to yield useful results from a
clustering approach. Since, as we will see, clustering does yield
benefits in this domain, it seems very plausible that it will also do
so for other, more naturally clustered domains.

\section{3. Clustering Algorithms}

There are many different criteria for quantifying the (dis)similarity
between (analyses of) two sentences or between two clusters of
sentences; Everitt (1993) provides a good overview. Unfortunately,
whatever the criterion selected, it is in general impractical to find
the optimal clustering of the data; instead, one of a variety of
algorithms must be used to find a locally optimal solution.

Let us for the moment consider the case where the language model
consists only of a unigram probability distribution for the words in
the vocabulary, with no $N$-gram (for $N>1$) or fuller linguistic
constraints considered. Perhaps the most obvious measure of the
similarity between two sentences or clusters is then Jaccard's
coefficient (Everitt, 1993, p41), the ratio of the number of words
occurring in both sentences to the number occurring in either or both.
Another possibility would be Euclidean distance, with each word in the
vocabulary defining a dimension in a vector space. However, it makes
sense to choose as a similarity measure the quantity we would like the
final clustering arrangement to minimize: the expected entropy (or,
equivalently, perplexity) of sentences from the domain. This goal is
analogous to that used in the work described earlier on finding word
classes by clustering.

For our simple unigram language model without clustering, the training
corpus perplexity is minimized (and its likelihood is maximized) by
assigning each word $w_i$ a probability $p_i = f_i/N$, where $f_i$ is
the frequency of $w_i$ and $N$ is the total size of the corpus. The
corpus likelihood is then $P_1 = \prod_{i}p_i^{f_i}$, and the
per-word entropy, $- \sum_{w_i} p_i log(p_i)$, is thus minimized. (See
e.g.\ Cover and Thomas, 1991, chapter 2 for the reasoning behind
this).

If now we model the language as consisting of sentences drawn at
random from $K$ different subpopulations, each with its own unigram
probability distribution for words, then the estimated corpus
probability is
\begin{center}
\begin{math}
P_K = \prod_{u_j} \sum_{c_k} q_k \prod_{w_i \in u_j} p_{k,i}
\end{math}
\end{center}
where the iterations are over each utterance $u_j$ in the corpus, each
cluster $c_1 \ldots c_K$ from which $u_j$ might arise, and each word
$w_i$ in utterance $u_j$.  $q_k = |c_k|/\sum_i |c_i|$ is the likelihood of an
utterance arising from cluster (or subpopulation) $c_k$, and $p_{k,i}$
is the likelihood assigned to word $w_i$ by cluster $k$, i.e.\ its
relative frequency in that cluster.

Our ideal, then, is the set of clusters that maximizes the
cluster-dependent corpus likelihood $P_K$. As with nearly all
clustering problems, finding a global maximum is impractical. To
derive a good approximation to it, therefore, we adopt the following
algorithm.
\begin{itemize}
\item Select a random ordering of the training corpus, and initialize
each cluster $c_k, k = 1 \ldots K$, to contain just the $k$th sentence
in the ordering.

\item Present each remaining training corpus sentence in turn,
initially creating an additional singleton cluster $c_{K+1}$ for it.
Merge that pair of clusters $c_1 \ldots c_{K+1}$ that entails the least
additional cost, i.e.\ the smallest reduction in the value of $P_K$
for the subcorpus seen so far.

\item When all training utterances have been incorporated, find all
the triples $(u,c_i,c_j),$ $i \neq j$, such that $u \in c_i$ but the
probability of $u$ is maximized by $c_j$. Move all such $u$'s (in
parallel) between clusters. Repeat until no further movements are
required.
\end{itemize}

In practice, we keep track not of $P_K$ but of the overall corpus
entropy $H_K = -log(P_K)$. We record the contribution each cluster
$c_k$ makes to $H_K$ as
\begin{center}
\begin{math}
H_K(c_k) = - \sum_{w_i \in c_k} f_{ik}log(f_{ik}/F_k)
\end{math}
\end{center}
where $f_{ik}$ is the frequency of $w_i$ in $c_k$ and $F_k$ =
$\sum_{w_j \in c_k} f_{jk}$, and find the value of this quantity for
all possible merged clusters.  The merge in the second step of the
algorithm is chosen to be the one minimizing the increase in entropy
between the unmerged and the merged clusters.

The adjustment process in the third step of the algorithm does not
attempt directly to decrease entropy but to achieve a clustering with
the obviously desirable property that each training sentence is best
predicted by the cluster it belongs to rather than by another cluster.
This heightens the similarities within clusters and the differences
between them. It also reduces the arbitrariness introduced into the
clustering process by the order in which the training sentences are
presented.\longfootnote{(Footnotes in this paper are used for the results
of statistical significance tests and other technical details not
essential to an understanding of the main argument).

This clustering algorithm is closely related to that of Ney,
Essen and Kneser (1994), who cluster words into equivalence classes
rather than training sentences into subcorpora. Ney {\it et al} begin
with a clustering in which the $K-1$ most frequent words each occupy a
singleton cluster and the $K$th cluster contains all the other words.
They then move words between clusters to maximize probabilities.  They
remark that ``other initialization schemes were found to work as well
and not to affect much the final result; however, their speed of
convergence may be much slower''. A frequency-based initialization
scheme of this kind is, however, less appropriate for clustering
sentences, because whereas very frequent words are likely to have
different distributions (the basis for Ney {\it et al}'s clustering),
some very frequent sentences may contain very similar word sequences
(the basis for ours), and it is therefore undesirable automatically to
put them in different clusters. In the ATIS corpus used for the
experiments described in Section \ref{expt}, for example, the first
and third most common sentence patterns are ``Show me the flights from
(city) to (city)'' and ``Show me flights from (city) to (city)''; and
our algorithm assigned these to the same cluster for 80\% of the runs
with ten or fewer clusters.

Iyer {\it et al} (1994) cluster training corpus paragraphs
agglomeratively on the basis of the proportion of content words in
common. This criterion is not related in any obvious way to perplexity
minimization, and would certainly be too blunt an instrument for clustering
ATIS sentences, which are fairly short and more limited in
vocabulary.}
The approach is applicable with only a minor modification to
$N$-grams for $N > 1$: the probability of a word within a cluster is
conditioned on the occurrence of the $N-1$ words preceding it, and the
entropy calculations take this into account.  Other cases of context
dependence modeled by a knowledge source can be handled similarly. And
there is no reason why the items characterizing the sentence have to
be (sequences of) words; occurrences of grammar rules, either without
any context or in the context of, say, the rules occurring just above
them in the parse tree, can be treated in just the same way.

\section{4. Experimental Results}
\label{expt}

Experiments were carried out to assess the effectiveness of
clustering, and therefore the existence of unexploited contextual
dependencies, for instances of two general types of language model.
In the first experiment, sentence hypotheses were evaluated on the
$N$-grams of words and word classes they contained.  In the second
experiment, evaluation was on the basis of grammar rules used rather
than word occurrences.

\subsection{$N$-gram Experiment}

In the first experiment, reference versions of a set of 5,873
domain-relevant (classes A and D) ATIS-2 sentences were allocated to
$K$ clusters for $K = 2, 3, 5, 6, 10$ and $20$ for the unigram, bigram
and trigram conditions and, for unigrams and bigrams only, $K=40$ and
$100$ as well. Each run was repeated for ten different random orders
for presentation of the training data. The unclustered ($K=1$) version
of each language model was also evaluated.  Some words, and some
sequences of words such as ``San Francisco'', were replaced by class
names to improve performance. \iflong{The per-item entropy of the training set
(i.e.\ the per-word entropy, but ignoring the need to distinguish
different words in the same class) was 6.04 for a unigram language
model, 2.96 for bigrams, and 1.97 for trigrams, giving perplexities of
65.7, 7.76 and 3.92 respectively. The greater the value of $K$, the
more a clustering reduced the apparent training set per-item entropy
(which, of course, is not the same thing as reducing test set
entropy). The reductions for $K=20$ were around 20\% for unigrams,
40\% for bigrams and 50\% for trigrams, with very little variation
(typically 1\% or less) between different runs for the same condition.}

The improvement (if any) due to clustering was measured by using the
various language models to make selections from N-best sentence
hypothesis lists; this choice of test was made for convenience rather
than out of any commitment to the N-best paradigm, and the techniques
described here could equally well be used with other forms of
speech-language interface.

Specifically, each clustering was tested against 1,354 hypothesis
lists output by a version of the DECIPHER (TM) speech recognizer
(Murveit {\it et al}, 1993) that itself used a (rather simpler) bigram
model. Where more then ten hypothesis were output for a sentence, only
the top ten were considered.  These 1,354 lists were the subset of two
1,000 sentence sets (the February and November 1992 ATIS evaluation
sets) for which the reference sentence itself occurred in the top ten
hypotheses. The clustered language model was used to select the most
likely hypothesis from the list without paying any attention either to
the score that DECIPHER assigned to each hypothesis on the basis of
acoustic information or its own bigram model, or to the ordering of
the list. In a real system, the DECIPHER scores would of course be
taken into account, but they were ignored here in order to maximize
the discriminatory power of the test in the presence of only a few
thousand test utterances.

To avoid penalizing longer hypotheses, the probabilities assigned to
hypotheses were normalized by sentence length. The probability
assigned by a cluster to an $N$-gram was taken to be the simple
maximum likelihood (relative frequency) value where this was non-zero.
When an $N$-gram in the test data had not been observed at all in the
training sentences assigned to a given cluster, a ``failure'',
representing a vanishingly small probability, was
assigned.\longfootnote{Failures, like log probabilities, were added
together; a derived sentence log probability therefore consisted of a
sum of log probabilities of the usual kind combined with a failure
count, i.e. a pair $(LP,F)$.  A difference in failure counts was
viewed as more significant than any difference in log probabilities;
formally,
\begin{center}
\begin{math}
(LP_1,F_1) < (LP_2,F_2) \Leftrightarrow F_1 > F_2 \vee (F_1 = F_2
\wedge LP_1 < LP_2).
\end{math}
\end{center}
Probabilities arising from different clusters were added as follows:
\begin{center}
\begin{tabular}{lll}
$(P_1,F_1) + (P_2,F_2) = $ & $(P_1,F_1)$ & if $F_1 < F_2$; \\
                           & $(P_2,F_2)$ & if $F_1 > F_2$; \\
                           & $(P_1+P_2,F_1)$ & if $F_1 = F_2.$ \\
\end{tabular}
\end{center}
where $P_i = e^{LP_i}$.

Although this scheme is quite adequate for hypothesis selection,
it means that no figures can be calculated for test set entropy analogous
to those for training set entropy.}
A number of backoff schemes of various degrees of sophistication,
including that of Katz (1987), were tried, but none produced any
improvement in performance, and several actually worsened it.

The average percentages of sentences correctly identified by clusterings
for each condition were as given in Table \ref{maintable}. The maximum
possible score was 100\%; the baseline score, that expected from a
random choice of a sentence from each list, was 11.4\%.

\begin{table}
\begin{center}
\begin{tabular}{|c|c|c|c|} \hline
Clusters & Unigram & Bigram & Trigram \\ \hline
1  & 12.4 & 34.3 & 51.6 \\ \hline
2  & 13.8 & 37.9 & 51.0 \\ \hline
3  & 15.3 & 39.5 & 50.8 \\ \hline
4  & 16.1 & 41.2 & 50.4 \\ \hline
5  & 16.8 & 41.2 & 51.0 \\ \hline
6  & 17.2 & 41.8 & 50.7 \\ \hline
10 & 17.8 & 43.1 & 51.2 \\ \hline
20 & 19.9 & 43.9 & 50.3 \\ \hline
40 & 22.3 & 45.0 &      \\ \hline
100& 24.4 & 46.4 &      \\ \hline
\end{tabular}
\caption{Average percentage scores for cluster-based $N$-gram models}
\label{maintable}
\end{center}
\end{table}
The unigram and bigram scores show a steady and, in fact,
statistically significant\shortfootnote{Details of significance tests
are omitted for space reasons. They are included in a longer version
of this paper available on request from the author.}\longfootnote{For
both unigrams and bigrams, the performance of the unclustered case was
compared with each of the ten runs for each clustered condition. All
the clustered runs scored better than the corresponding unclustered
case.  For each clustered run, the McNemar change test (Siegel and
Castellan, 1988, p75) was applied to the number of sentences for which
an improvement was observed (incorrect choice by unclustered model,
correct by clustered model) and the number with a deterioration
(correct by unclustered, incorrect by clustered). At the P=0.05 level,
two-tail, the clustered unigram results were significantly better for
three of the ten two-cluster runs, nine of the ten three-cluster runs,
and all the runs for more than three clusters. For the bigram case,
all runs except one of the two-cluster ones were significantly better
than the unclustered result.

The Wilcoxon-Mann-Whitney test ({\it ibid}, p128) was applied to the
scores for all ten runs at each of two numbers of clusters $K1$ and
$K2$. It showed that for unigrams, for all $K1>K2>1$, having $K1$
clusters is significantly better at the P=0.05 level than having $K2$
except for the cases $(K1,K2)=(5,4)$, $(6,5)$ and $(10,6)$, where no
significant difference was found. For bigrams, the difference between
4, 5 or 6 clusters was not significant, and neither was that between
10 and 20, but all other $(K1,K2)$ pairs were significantly
different.} increase with the number of clusters.  Using twenty
clusters for bigrams (score 43.9\%) in fact gives more than half the
advantage over unclustered bigrams that is given by moving from
unclustered bigrams to unclustered trigrams.  However, clustering
trigrams produces no improvement in score; in fact, it gives a small
but statistically significant\longfootnote{The McNemar test revealed no
cases of clustered trigrams performing significantly better than
unclustered.  However, in 15 of the 70 runs, the performance was
significantly worse. The Wilcoxon-Mann-Whitney test showed no clear
advantage or disadvantage in different numbers of clusters for
trigrams.} deterioration, presumably due to the increase in the number
of parameters that need to be calculated.

The random choice of a presentation order for the data meant that
different clusterings were arrived at on each run for a given
condition ($(N,K)$ for $N$-grams and $K$ clusters). There was some
limited evidence that some clusterings for the same condition were
significantly better than others, rather than just happening to
perform better on the particular test data used.\longfootnote{A positive
correlation was found between scores on the February 1992 and November
1992 parts of the test set for eight of the nine unigram conditions;
the correlation was positive and significant (P=0.05 level, two-tail)
for two of these conditions. For bigrams, the correlation was less
clear, and for trigrams, not apparent at all.} More trials would be
needed to establish whether presentation order does in general make a
genuine difference to the quality of a clustering. If there is one,
however, it would appear to be fairly small compared to the
improvements available (in the unigram and bigram cases) from
increasing the numbers of clusters.

\subsection{Grammar Rule Experiment}

In the second experiment, each training sentence and each test
sentence hypothesis was analysed by the Core Language Engine (Alshawi,
1992) trained on the ATIS domain (Agn\"as {\it et al}, 1994).
Unanalysable sentences were discarded, as were sentences of over 15
words in length (the ATIS adaptation had concentrated on sentences of
15 words or under, and analysis of longer sentences was less reliable
and slower). When a sentence was analysed successfully, several
semantic analyses were, in general, created, and a selection was made
from among these on the basis of trained preference functions (Alshawi
and Carter, 1994). For the purpose of the experiment, clustering
and hypothesis selection were performed on the basis not of the words
in a sentence but of the grammar rules used to construct its most
preferred analysis.

The simplest condition, hereafter referred to as ``1-rule'', was
analogous to the unigram case for word-based evaluation. A sentence
was modeled simply as a bag of rules, and no attempt (other than the
clustering itself) was made to account for dependencies between rules.

Another condition, henceforth ``2-rule'' because of its analogy to
bigrams, was also tried. Here, each rule occurrence was represented
not in isolation but in the context of the rule immediately above it
in the parse tree\iflong{(its ``predecessor'' if the tree is traversed
top-down). This choice was made on the assumption that the immediately
dominating rule would be one important influence on the likelihood of
occurrence of a particular rule. Other choices, involving sister rules
and/or rules in less closely related positions, or the compilation of
rules into common combinations (Samuelsson and Rayner,
1991)}\ifshort{. Other choices of context} might have worked as well
or better; our purpose here is simply to illustrate and assess ways in
which explicit context modeling can be combined with clustering.

The training corpus consisted of the 4,279 sentences in the
5,873-sentence set that were analysable and consisted of fifteen words
or less. The test corpus consisted of 1,106 hypothesis lists, selected
in the same way (on the basis of length and analysability of their
reference sentences) from the 1,354 used in the first experiment. The
``baseline'' score for this test corpus, expected from a random choice
of (analysable) hypothesis, was 23.2\%. This was rather higher than the
11.4\% for word-based selection because the hypothesis lists used were
in general shorter, unanalysable hypotheses having been excluded.

The average percentages of correct hypotheses (actual word strings,
not just the rules used to represent them) selected by the 1-rule and
2-rule conditions were as given in Table \ref{ruletable}.
\begin{table}
\begin{center}
\begin{tabular}{|c|c|c|} \hline
Clusters & 1-rule & 2-rule \\ \hline
1  & 29.4 & 34.3 \\ \hline
2  & 31.4 & 35.5 \\ \hline
3  & 31.8 & 36.2 \\ \hline
4  & 31.7 & 37.0 \\ \hline
5  & 32.3 & 37.2 \\ \hline
6  & 31.9 & 37.3 \\ \hline
10 & 32.8 & 37.5 \\ \hline
20 & 35.1 & 38.3 \\ \hline
40 & 35.8 & 38.9 \\ \hline
\end{tabular}
\caption{Average percentage scores for cluster-based $N$-rule models}
\label{ruletable}
\end{center}
\end{table}

These results show that clustering gives a significant advantage for
both the 1-rule and the 2-rule types of model,\longfootnote{For both
the 1-rule and the 2-rule condition, nearly all the runs for all
values of $K$ were significantly better than the unclustered case: all
the clustered scores were higher than the corresponding unclustered
one, and the difference was significant under the McNemar test in 131
of the 140 cases.} and that the more clusters are created, the larger
the advantage is, at least up to $K=20$ clusters.\longfootnote{The
Wilcoxon-Mann-Whitney test suggested that, for the 1-rule condition,
clusterings with $K=20$ and $K=40$ clusters were significantly better
than all other values of $K$ tried, but not significantly different
from either other. $K=2$ was significantly worse than all larger $K$
values except $K$=4. Most other comparisons were not significant.  For
the 2-rule case, $K=2$ was significantly worse than $K=3$, which in
turn was significantly worse than all $K>3$. $K=10$ and $K=20$ did not
differ significantly, but apart from that, $K=20$ and $K=40$ were
again significantly better than all other $K$, but did not differ
significantly from each other.} As with the $N$-gram experiment, there
is weak evidence that some clusterings are genuinely better than
others for the same condition.\longfootnote{A significant positive
correlation was found between scores on the February 1992 and November
1992 parts of the test set for three of the eight 1-rule conditions;
however, one significant (but presumably coincidental) {\it negative}
correlation was also found.  For 2-rule conditions, one significant
positive correlation and no significant negative ones were found.}

\section{5. Conclusions}

I have suggested that training corpus clustering can be used both to
extend the effectiveness of a very general class of language models,
and to provide evidence of whether a particular language model could
benefit from extending it by hand to allow it to take better
account of context. Clustering can be useful even when there is no
reason to believe the training corpus naturally divides into any
particular number of clusters on any extrinsic grounds.

The experimental results presented show that clustering increases the
(absolute) success rate of unigram and bigram language modeling for a
particular ATIS task by up to about 12\%, and that performance
improves steadily as the number of clusters climbs towards 100
(probably a reasonable upper limit, given that there are only a few
thousand training sentences). However, clusters do not improve trigram
modeling at all. This is consistent with experience (Rayner {\it et
al}, 1994) that, for the ATIS domain, trigrams model inter-word
effects much better than bigrams do, but that extending the $N$-gram
model beyond $N=3$ is much less beneficial.

For $N$-rule modeling, clustering increases the success rate for both
$N=1$ and $N=2$, although only by about half as much as for $N$-grams.
This suggests that conditioning the occurrence of a grammar rule on
the identity of its mother (as in the 2-rule case) accounts for some,
but not all, of the contextual influences that operate. From this it
is sensible to conclude, consistently with the results of Briscoe and
Carroll (1993), that a more complex model of grammar rule interaction
might yield better results. Either conditioning on other parts of the
parse tree than the mother node could be included, or a rather
different scheme such as Briscoe and Carroll's could be used.

Neither the observation that trigrams may represent the limit of
usefulness for $N$-gram modeling in ATIS, nor that non-trivial
contextual influences exist between occurrences of grammar rules, is
very novel or remarkable in its own right. Rather, what is of interest
is that the improvement (or otherwise) in particular language models
from the application of clustering is consistent with those
observations. This is important evidence for the main hypothesis of
this paper: that enhancing a language model with clustering, which
once the software is in place can be done largely automatically, can
give us important clues about whether it is worth expending research,
programming, data-collection and machine resources on hand-coded
improvements to the way in which the language model in question models
context, or whether those resources are best devoted to different,
additional kinds of language model.

\section*{Acknowledgements}

This research was partly funded by the Defence Research Agency,
Malvern, UK, under assignment M85T51XX.

I am grateful to Manny Rayner and Ian Lewin for useful
comments on earlier versions of this paper. Responsibility for any
remaining errors or unclarities rests in the customary place.

\iflong{A shorter version of this paper appears in
the Proceedings of the ACL Conference on Applied Natural Language
Processing, Stuttgart, October 1994, and is \copyright{} Association
for Computational Linguistics.}

\section*{References}

\newenvironment{reverseindent}%
{\begin{list}{}{\setlength{\labelsep}{0in}
                \setlength{\labelwidth}{0in}
                \setlength{\itemsep}{0in}
                \setlength{\itemindent}{-\leftmargin}}}%
{\end{list}}

\begin{reverseindent}


\item
Agn\"{a}s, M-S., {\it et al} (1994). {\it Spoken Language Translator
First Year Report}.  SRI International Cambridge Technical Report
CRC-043.

\item
Alshawi, H., and D.M.~Carter (1994). ``Training and Scaling Preference
Functions for Disambiguation''. {\it Computational Linguistics} (to
appear).

\item
Briscoe, T., and J.~Carroll (1993). ``Generalized Probabilistic LR
Parsing of Natural Language (Corpora) with Unification-Based
Grammars'', {\it Computational Linguistics}, Vol 19:1, 25-60.

\item
Cover, T.M., and J.A.~Thomas (1991). {\it Elements of Information
Theory}. New York: Wiley.

\item
Everitt, B.S.~(1993). {\it Cluster Analysis}, Third Edition. London:
Edward Arnold.

\item
Iyer, R., M.~Ostendorf and J.R.~Rohlicek (1994). ``Language Modeling
with Sentence-Level Mixtures''. {\it Proceedings of the ARPA Workshop
on Human Language Technology}.

\item
Jelinek, F., B.~Merialdo, S.~Roukos and M.~Strauss (1991). ``A Dynamic
Language Model for Speech Recognition'', {\it Proceedings of the
Speech and Natural Language DARPA Workshop}, Feb 1991, 293-295.

\item
Katz, S.M.~(1987). ``Estimation of Probabilities from Sparse Data for
the Language Model Component of a Speech Recognizer'', {\it IEEE
Transactions on Acoustics, Speech and Signal Processing}, Vol
ASSP-35:3.

\item
Lewin, I., D.M.~Carter, S.~Pulman, S.~Browning, K.~Ponting and
M.~Russell (1993). ``A Speech-Based Route Enquiry System Built From
General-Purpose Components'', {\it Proceedings of Eurospeech-93}.

\item
Murveit, H., J.~Butzberger, V.~Digalakis and M.~Weintraub (1993).
``Large Vocabulary Dictation using SRI's DECIPHER(TM)
Speech Recognition System: Progressive Search Techniques'',
{\it Proceedings of the International Conference on
Acoustics, Speech and Signal Processing}, Minneapolis, Minnesota.

\item
Ney, H., U.~Essen and R.~Kneser (1994). ``On Structuring Probabilistic
Dependencies in Stochastic Language Modeling''. {\it Computer Speech
and Language}, vol 8:1, 1-38.

\item
Pereira, F., N.~Tishby and L.~Lee (1993). ``Distributional Clustering
of English Words''. {\it Proceedings of ACL-93}, 183-190.

\item
Rayner, M., D.~Carter, V.~Digalakis and P.~Price (1994).  ``Combining
Knowledge Sources to Reorder N-best Speech Hypothesis Lists''. {\it
Proceedings of the ARPA Workshop on Human Language Technology}.

\item
Rosenfeld, R. (1994). ``A Hybrid Approach to Adaptive Statistical
Language Modeling''. {\it Proceedings of the ARPA Workshop on Human
Language Technology}.

\iflong{
\item
Samuelsson, C., and M.~Rayner (1991). ``Quantitative Evaluation of
Explanation-Based Learning as a Tuning Tool for a Large-Scale Natural
Language System''.  {\it Proceedings of 12th International Joint
Conference on Artificial Intelligence}.  Sydney, Australia.

\item
Siegel, S., and N.J.~Castellan (1988). {\it Nonparametric Statistics},
Second Edition. New York: McGraw-Hill.}

\end{reverseindent}

\end{document}